\def\printing{myijcai}
\def\myijcai{myijcai}
\def\blind{blind}
\def\article{article}
\def\frontpage{frontpage}
\ifx\printing\myijcai\documentclass{article} \usepackage{ijcai03}
   \else\ifx\printing\blind\documentclass{article} \usepackage{ijcai03}\else
   \documentclass{article}\fi\fi
\usepackage{times}
\usepackage{epic}
\usepackage{ecltree}
\usepackage{graphicx}
\usepackage{latexsym}
\newcommand{\la}{\langle}
\newcommand{\ra}{\rangle}

\newcommand{\And}{\wedge}

\newenvironment{Rem}[1]{\leavebigvspace\noindent
{\tt ** #1 **}\newline}{$\phantom{a}$\newline{\tt ********}\leavebigvspace}

\def\halmos{\mbox{\hspace*{1pc}$\Box$}}

\newcommand{\leavebigvspace}{\par \addvspace{\bigskipamount}}
\def\theexample{\thesection.\arabic{example}}
\newtheorem{Example}{Example}[section]
{{$\phantom{a}$%
\leavebigvspace}%
\end{Example}}
\def\thetheorem{\thesection.\arabic{theorem}}
\newtheorem{Theorem}{Theorem}[section]
{{$\phantom{a}$}\end{Theorem}}
\def\thedefinition{\thesection.\arabic{definition}}
\newtheorem{Definition}{Definition}[section]
{{$\phantom{a}$}\end{Definition}}

\begin{document}

\newlength{\halftextwidth}
\setlength{\halftextwidth}{0.47\textwidth}
\def\halffigsize{2.2in}
%\newlength{\thirdtextwidth}
%\setlength{\thirdtextwidth}{0.31\textwidth}
\def\thirdfigsize{1.5in}
\def\negvspace{0in}
\def\posvspace{0em}

\input epsf

%\usepackage{amssymb}

%\addtolength{\textwidth}{1.5in}
%\addtolength{\textwidth}{0.5in}
%\addtolength{\oddsidemargin}{-0.75in}
%\addtolength{\topmargin}{-0.5in}
%\addtolength{\textheight}{0.5in}

%\newcommand{\figref}[1]{Figure \ref{#1}}
%\newcommand{\tblref}[1]{Table \ref{#1}}
%\newcommand{\secref}[1]{Section \ref{#1}}
%\newcommand{\eqref}[1]{(\ref{#1})}
%\newcommand{\mygtrsim}{\gtrsim}

\newcommand{\deriv}[2]{\Delta #1_#2}

\newcommand{\tighter}[2]{\mbox{$#1 \preceq #2$}}
\newcommand{\stighter}[2]{\mbox{$#1 \prec #2$}}
\newcommand{\ac}[1]{\mbox{$ac(#1)$}}
\newcommand{\acmin}[1]{\mbox{$ac_{\min}(#1)$}}
\newcommand{\pc}[1]{\mbox{$pc(#1)$}}
\newcommand{\spc}[1]{\mbox{$spc(#1)$}}
\newcommand{\incomparable}[2]{\mbox{$#1 \sim #2$}}
\newcommand{\twodarray}[4]{\mbox{\scriptsize $\left( \hspace{-0.5em} \begin{array}{cc} #1 & #2 \\ #3 & #4 \end{array} \hspace{-0.5em} \right)$}}

\newcommand{\threedarray}[9]{\mbox{\scriptsize $\left( \hspace{-0.5em} \begin{array}{ccc} #1 & #2 & #3 \\ #4 & #5 & #6 \\ #7 & #8 & #9  \end{array} \hspace{-0.5em} \right)$}}

\renewcommand{\theenumii}{\alph{enumii}}
\renewcommand{\theenumiii}{\roman{enumiii}}
\newcommand{\figref}[1]{Figure \ref{#1}}
\newcommand{\tref}[1]{Table \ref{#1}}
\newcommand{\size}{\mbox{$N$}} %\newcommand{\prob}[1]{\mbox{Pr\{#1\}}}
\newcommand{\prob}[1]{\mbox{{P}r\{\scriptsize#1\}}}
\newcommand{\probB}[1]{\mbox{{P}r\{#1\}}}
\newcommand{\secref}[1]{Section \ref{#1}}
\newcommand{\myfrac}[2]{(#1)/#2}
\newcommand{\CSP}{\mbox{\sc Csp}}
\newcommand{\Csp}{\mbox{\sc Csp}}
\newcommand{\SAT}{\mbox{\sc Sat}}

\def\op{\mbox{$\kappa$}}
\def\opcrit{\mbox{$\kappa_{c}$}}
\def\opn{\mbox{$\gamma$}}
\newcommand{\nmpp}{\mbox{$\langle n,m,p_{1},p_{2}\rangle$}}
\newtheorem{mydefinition}{Definition}
\newtheorem{mytheorem}{Theorem}
\newtheorem{mytheorem1}{Theorem}
\newcommand{\myproof}{\noindent {\bf Proof:\ \ }}
\newcommand{\myqed}{\mbox{QED.}}
\newcommand{\myAnd}{\ \& \ }
\newcommand{\myOr}{\ \vee \ }
\newcommand{\myNot}{\neg}
\newcommand{\msetl}{\{ \hspace{-0.3em} \{}
\newcommand{\msetr}{\} \hspace{-0.3em} \}}
\newcommand{\mymin}[1]{min(#1)}
\newcommand{\mymax}[1]{max(#1)}
\newcommand{\myfloor}[1]{\mbox{\it floor}(#1)}
\newcommand{\myceiling}[1]{\mbox{\it ceil}(#1)}
\newcommand{\myhspaceOne}{\hspace{0.5em}}
\newcommand{\myhspaceTwo}{\hspace{1em}}
\newcommand{\myhspaceThree}{\hspace{1.5em}}
\newcommand{\myhspaceFour}{\hspace{2em}}
\newcommand{\myhspaceFive}{\hspace{2.5em}}

\title{Multiset Ordering Constraints}
\ifx\printing\blind
\author{%Toby Walsh\\
%Department of Computer Science \\
%University of York \\
%York YO10 5DD  \\
%England \\
%{\tt tw@cs.york.ac.uk}}
Content areas: constraint programming, constraint satisfaction\\
}
\date{}
\else \fi

\ifx\printing\myijcai
\author{Alan Frisch \\
Dept of CS \\
University of York\\
England \\
frisch@cs.york.ac.uk \And
Brahim Hnich\\
4C Lab \\
University College Cork\\
Ireland\\
b.hnich@4c.ucc.ie \And
Zeynep Kiziltan\\
CS Division\\
Uppsala University\\
Sweden\\
zeynep@4c.ucc.ie \And
Ian Miguel \\
Dept of CS\\
University of York\\
England\\
ianm@cs.york.ac.uk \And
Toby Walsh\\
4C Lab\\
University College Cork\\
Ireland\\
tw@4c.ucc.ie
}
%\date{}
%\setlength\titlebox{2in}
\else \fi

\ifx\printing\article
\author{Alan Frisch \\
Dept of CS \\
University of York\\
England \\
frisch@cs.york.ac.uk \And
Brahim Hnich\\
4C Lab \\
University College Cork\\
Ireland\\
b.hnich@4c.ucc.ie \And
Zeynep Kiziltan\\
CS Division\\
Uppsala University\\
Sweden\\
zeynep@4c.ucc.ie \And
Ian Miguel \\
Dept of CS\\
University of York\\
England\\
ianm@cs.york.ac.uk \And
Toby Walsh\\
4C Lab\\
University College Cork\\
Ireland\\
tw@4c.ucc.ie
}
\date{1st Jan 2003}
\else \fi

\ifx\printing\frontpage
\author{
Alan Frisch \\
Dept of CS \\
University of York\\
England \\
frisch@cs.york.ac.uk \and
Brahim Hnich\\
4C Lab \\
University College Cork\\
Ireland\\
b.hnich@4c.ucc.ie \and
Zeynep Kiziltan\\
CS Division\\
Uppsala University\\
Sweden\\
zeynep@4c.ucc.ie \and
Ian Miguel \\
Dept of CS\\
University of York\\
England\\
ianm@cs.york.ac.uk \and
Toby Walsh\\
4C Lab\\
University College Cork\\
Ireland\\
tw@4c.ucc.ie }
 \else \fi

\maketitle
\begin{abstract}
We identify a new and important global (or non-binary) constraint. This
constraint ensures that the values taken by two vectors of variables,
when viewed as multisets, are ordered.  This constraint is useful for
a number of different applications including breaking symmetry and
fuzzy constraint satisfaction.  We propose and implement an efficient
linear time algorithm for enforcing generalised arc-consistency on
such a multiset ordering constraint.
%We have implemented this
%algorithm in ILOG's Solver toolkit.
Experimental
results on several problem
domains show considerable promise.
\end{abstract}

\ifx\printing\frontpage
\begin{center}
\begin{tabular}{ll}
Content areas: constraint programming, constraint satisfaction\\
%Word count: & 4668 words \\
%Tracking number: & 142 \\
\end{tabular}
\end{center}

\begin{quote}
This paper has not already been accepted by and is not currently under review for
a journal or another conference, nor will it be submitted for such during IJCAI's
review period.
\end{quote}

\else \fi

\ifx\printing\frontpage
\eject \end{document} \else \fi

%\vspace{-2mm}
\section{Introduction}

Global (or non-binary) constraints are one of the factors central to
the success of constraint programming
% ZEYNEP
%\cite{beldiceanu2,regin1,regin2,Bessiere-Regin97,regin4,fhkmwcp2002}.
\cite{regin1,regin2,beldiceanu1}.
Global constraints specify patterns that occur in many problems, and
call efficient and effective constraint propagation
algorithms. 
% For example, we often have sets of variables which must
%take different values (e.g. activities in a scheduling problem
%requiring the same resource must all be assigned different
%times). Most constraint solvers therefore provide a global
%all-different constraint which is propagated efficiently
%algorithm %due to Regin for propagating this constraint effectively
%\cite{regin1}.
%See
%\cite{beldiceanu1} for other global constraints.
%
In this paper, we identify a new and important global constraint.
This constraint
ensures that the values taken by two vectors of variables, when viewed
as multisets, are ordered. Such a constraint is useful in a number of
domains. For example,
%in the template design problem (prob002 in
%CSPLib: www.csplib.org)
%ZEYNEP
%\cite{csplib})
%, we wish to assign designs to
%printing templates. As there are a fixed number of slots for
%designs on each template, we can model this with a vector of
%variables for each template. Each variable is assigned the design
%for a particular slot. This model has an unnecessary symmetry as
%we can freely permute slots and templates. We can break this
%symmetry by considering the multiset of values associated with
%each vector and ordering these multisets. Note that it is a
%multiset of values as designs may be repeated, and the order of
%the designs is unimportant.
%
%As a second example,
in the progressive party problem (prob013 in csplib.org), we wish
to assign a host for each guest and period. We can model this with
a vector of variables for each period. Each variable is assigned
the host for a particular guest. This model has unnecessary
symmetry as the periods can be freely permuted. We
can break this symmetry by considering the multiset of values
associated with each vector and ordering these multisets.% It
%is a multiset as several guests can have the same host.
%In both examples, we viewed the values
%taken by a vector of variables
%as a multiset and broke a symmetry
%by posting an ordering
%constraint between these multisets.
The aim of this paper is to study such
multiset ordering constraints and to
develop efficient and effective techniques
for enforcing them.

% ====================================================================
\section{Formal Background}
% ====================================================================
A constraint satisfaction problem (CSP) consists of a set of
variables, each with a finite domain of values, and a set of
constraints that specify allowed values for subsets of variables.  A
solution is an assignment of values to the variables satisfying the
constraints. To find such solutions, we %constraint solvers often
explore partial assignments enforcing a local consistency like
generalized arc-consistency (GAC). A constraint is GAC iff, when a
variable in the constraint is assigned a value, compatible values
exist for all the other variables in the constraint. GAC reduces to
arc-consistency (AC) for binary constraints and to node-consistency
(NC) for unary constraints. Finally, a constraint is bounds consistent
(BC) iff, when a variable in the constraint is assigned its maximum or
minimum value, there exist compatible values for all the other
variables in the constraint. If a constraint $c$ is NC, BC, AC or GAC
then we write NC($c$), BC($c$), AC($c$) or GAC($c$) respectively.

We now define the multiset ordering.
A multiset is an unordered list in which repetition is
allowed. %, written $\msetl x_0, \ldots , x_{n-1} \msetr $.
%We write $max(M)$ for the maximum element of a multiset $M$.
%This presuppose a total ordering
%on elements of the multiset. This in turn
%induces a total ordering on multisets. Formally,
A multiset $M$ is ordered smaller than another $N$, written
$M \prec_{\rm m} N$ iff
%either ($M=\msetl \msetr$ and $N \neq \msetl \msetr$)
%or $max(M) < max(N)$
%or ($max(M)=max(N)$ and $M-\msetl max(M) \msetr \prec_{\rm m} N-\msetl max(N) \msetr$).
%That is,
either $M$ is empty and $N$ is not,
or the largest value in $M$ is smaller than the largest
value in $N$, or the largest values are the same and, if
we eliminate one occurrence of the largest value from both $M$ and $N$,
the resulting two multisets are ordered.
We can weaken
the ordering to include multiset equality.
That is,  $M \preceq_{\rm m} N$ iff $M=N$ or $M \prec_{\rm m} N$.
As in the introductory example, we will often view the values
taken by a vector of variables as a multiset. Given two vectors, $\vec x$ and
$\vec y$, we write a multiset ordering constraint as $\vec x \leq_{\rm
m} \vec y$.  This constraint is satisfied iff the values for the
variables in the vectors, when treated as multisets, satisfy the
multiset ordering.  Similarly, we write a strict multiset ordering
constraint as $\vec x <_{\rm m} \vec y$.
Throughout the paper, we assume that the variables 
being ordered are disjoint and not repeated. 

We also need the following %additional
notation. Input vectors are indexed
from 0. The minimum element in the domain of $x_i$ is %denoted by
$\mymin{x_i}$, and the maximum is $\mymax{x_i}$.  The floor
function,
$\myfloor{\vec x}$ assigns all variables in %a vector
$\vec x$ to their
minimum values, whilst the ceiling function, $\myceiling{\vec x}$
assigns all to their maximums.
The vector $\vec{x}_{v=d}$ is
identical to $\vec{x}$ except $v$ now has the %singleton
domain $\{d\}$.
The function $occ(\vec x)$
computes the occurrence vector associated with $\vec x$. We index
occurrence vectors in decreasing order %of significance 
from the maximum to the minimum value from the domains in $\vec x$.
When
comparing two occurrence vectors, we assume they start %and end 
with
the occurrence of the same value, adding leading %and trailing 
zeroes
as necessary. Finally, $\vec x \leq_{lex} \vec y$ iff $\vec x$ is
lexicographically less than or equal to $\vec y$.

% ====================================================================
\section{Motivating applications}
% ====================================================================

% ====================================================================
\subsection{Matrix symmetry}

Many constraints programs contain matrices of decision variables (so
called ``matrix models''), and the rows and/or columns of these
matrices are symmetric and can be permuted \cite{ffhkmpwcp2002}. Such
symmetries are very difficult to deal with as there are a
super-exponential number of permutations of the rows or columns to
consider. There are several ways to break symmetry in a CSP, such as
SBDS \cite{gent:SBDS} or SBDD \cite{fahle:SBDD}. One of the most
effective, and the one which we will concentrate on as a major
application for a multiset ordering constraint, is adding extra
symmetry-breaking constraints to an initial model. Existing techniques
for dealing with such symmetries typically eliminate only some of the
symmetry. Additional techniques, like those proposed here, are
therefore of considerable value.

The progressive party problem mentioned earlier has a 2d matrix of
decision variables with matrix symmetry. The rows of the matrix
are the guests, the columns are the periods. Each variable gives
the host assigned to a given guest in a given period. As periods
are indistinguishable, the columns of the matrix are symmetric.
One way to break such column symmetry is to lex order the columns
\cite{fhkmwcp2002}. Similarly, as guests can be indistinguishable,
(some of) the rows may be symmetric and can be lex ordered.
Alternatively, we can treat each row and/or column as a multiset
and break such %row and/or column 
symmetry by multiset ordering the
rows and/or columns. 

Unlike lex ordering,
multiset ordering the rows of a matrix may
not eliminate all row symmetry. 
For example, consider the symmetric matrices:

\vspace{-0.5em}
{\scriptsize
$$
\left(
\begin{array}{cc}
0 & 0 \\
0 & 1
\end{array}
\right)
~ ~ ~ ~
\left(
\begin{array}{cc}
0 & 0 \\
1 & 0
\end{array}
\right)
$$
}

\vspace{-0.7em} \noindent
Both satisfy the constraint that the first
row is multiset less than the second.
It is therefore
a little surprising to discover that
multiset ordering (which does not break all row
symmetry) is not dominated by lex
ordering (which does) but is incomparable. 
For example, 
$\langle 0, 2 \rangle \leq_{lex} \langle 1, 1 \rangle$
but
$\langle 1, 1 \rangle \leq_{m} \langle 0, 2 \rangle$.

When we have both row and column symmetry, we can multiset order
both rows and columns. Like lex ordering both rows and columns,
this may not eliminate all row and column symmetry. 
Consider the symmetric matrices:

\vspace{-0.5em}
{\scriptsize
$$
\left(
\begin{array}{ccc}
0 & 1 & 1 \\
1 & 0 & 1
\end{array}
\right)
~ ~ ~ ~
\left(
\begin{array}{ccc}
1 & 0 & 1\\
0 & 1 & 1
\end{array}
\right)
$$
}

\vspace{-0.7em} \noindent
Both have multiset ordered rows and columns. Unsurprisingly, multiset
ordering rows and columns is incomparable to lex ordering rows and
columns. Consider the symmetric matrices:

\vspace{-0.5em}
{\scriptsize
$$
\left(
\begin{array}{ccc}
0 & 0 & 1 \\
0 & 1 & 0 \\
1 & 1 & 0
\end{array}
\right)
~ ~ ~ ~
\left(
\begin{array}{ccc}
0 & 1 & 0 \\
0 & 0 & 1 \\
1 & 0 & 1
\end{array}
\right)
$$
}

\vspace{-0.7em} \noindent
The first has lex ordered rows and columns, but the columns are not
multiset ordered.  The second has rows and columns that are multiset
ordered but the columns are not lex ordered.

An alternative way to deal with row and column symmetry
is to multiset order in one dimension
and apply the symmetry breaking method of our choice in the other dimension.
This is one of the best features of using multiset ordering to
break symmetry. It is compatible with any other method in the other
dimension. For instance, we can multiset order the
rows and lex order the columns. Preliminary results in
\cite{kscp02} suggest that such a combined method is very
promising. % on many problems.
This combined method  does not eliminate all 
symmetry (but it is unlikely that any polynomial
set of constraints does). 
Consider the symmetric matrices:

\vspace{-0.5em}
{\scriptsize
$$
\left(
\begin{array}{ccc}
0 & 1 & 1 \\
1 & 0 & 1 \\
1 & 2 & 0
\end{array}
\right)
~ ~ ~ ~
\left(
\begin{array}{ccc}
0 & 1 & 1\\
1 & 0 & 1\\
2 & 1 & 0
\end{array}
\right)
$$
}

\vspace{-0.7em} \noindent
Both have rows that are multiset ordered, and rows
and columns that are lex ordered.
Multiset ordering the rows and lex ordering
the columns is again incomparable to lex ordering
rows and columns.
Consider the symmetric matrices:

\vspace{-0.5em}
{\scriptsize
$$
\left(
\begin{array}{ccc}
0 & 0 & 1 \\
0 & 1 & 0 \\
1 & 0 & 0 \\
1 & 2 & 0 \\
0 & 2 & 2
\end{array}
\right)
~ ~ ~ ~
\left(
\begin{array}{ccc}
0 & 0 & 1 \\
0 & 1 & 0 \\
0 & 2 & 2 \\
1 & 0 & 0 \\
1 & 0 & 2
\end{array}
\right)
$$
}

\vspace{-0.7em} \noindent
The first matrix has rows that
are multiset ordered and columns
that are lex ordered.
However, its rows are not lex
ordered. The second matrix has rows and columns
that are lex ordered but does not have
rows that are multiset ordered.
Whilst the two orderings are theoretically incomparable,
our experimental results (see later) show that
multiset ordering the rows and lex ordering the
columns is often the most effective symmetry
breaking constraint currently known.

%  ZEYNEP

% ====================================================================
\subsection{Fuzzy constraints}

Another application for multiset ordering is to fuzzy CSPs.  A
fuzzy constraint associates a degree of satisfaction to an
assignment tuple for the variables it constrains. To combine
degrees of satisfaction, we can use a combination operator like
the minimum function. Unfortunately, the minimum function may
cause a {\em drowning effect} when one poorly satisfied constraint
`drowns' many highly satisfied constraints.  One solution is to
collect a vector of degrees of satisfaction, sort these values in
ascending order and compare them lexicographically.  This {\em
leximin} combination operator identifies the assignment that
violates the fewest constraints \cite{fargier:Thesis}.  This
induces an ordering identical to the multiset ordering except that
the lower elements of the satisfaction scale are the more
significant.  It is simple to modify a multiset ordering
constraint to consider the values in a reverse order.  To solve
such leximin fuzzy CSPs, we can then use branch and bound, adding
an ordering constraint when we find a solution to ensure that
future solutions are greater in the leximin ordering.
% than $\vec{\sigma}$.

% ====================================================================
\section{GAC algorithm for multiset ordering}
% ====================================================================

The last section motivated why we want %to post 
multiset ordering constraints. We need, however, to be able
to propagate such constraints efficiently.
We therefore developed an efficient
GAC algorithm for such constraints.

% ====================================================================
\subsection{Background}

The algorithm exploits two theoretical results. The first
reduces the problem to testing support for
upper bounds of $\vec x$ and lower
bounds of $\vec y$ on suitable ground vectors. The second reduces
these tests to lex ordering suitable occurrence vectors.
Identical results hold for the strict multiset ordering constraint
but for reasons of space we omit them here.

\noindent
{\bf Lemma 1}
{\em Given two disjoint and non-repeating vectors 
of variables, $\vec{x}$ and $\vec{y}$,
with non-empty domains, $GAC(\vec{x} \leq_m \vec{y})$
iff \,$\forall x_i\in\vec{x}, y_j\in\vec{y}$:}
{\small
\begin{eqnarray*}
\myfloor{\vec{x}_{x_i= \mymax{x_i}}} &\leq_{m}& \myceiling{\vec{y}} \\
\myfloor{\vec{x}} &\leq_{m}& \myceiling{\vec{y}_{y_j = \mymin{y_j}}}
\end{eqnarray*}
}
%\end{mytheorem}
\myproof 
($\Rightarrow$)
As the constraint is GAC, all values have support.
In particular, $x_i=\mymax{x_i}$ has support.
The best support comes if all the other variables in $\vec{x}$ 
take their minimums, and all the variables in 
$\vec{y}$ take their maximums. Hence,
$\myfloor{\vec{x}_{x_i= \mymax{x_i}}} \leq_{m} \myceiling{\vec{y}}$. 
Similarly, for $y_i$. 

($\Leftarrow$)
The first constraint
ensures that $\mymax{x_i}$ is supported.  The values which support
$\mymax{x_i}$ also support all values smaller.  Hence, all the values
in the domain of $x_i$ are supported. By an analogous argument, all
the values in the domain of $y_i$ are supported. Hence the
constraint is GAC. \myqed

The next lemma reduces
these test for support to lex ordering suitable occurrence vectors.

\noindent
{\bf Lemma 2} {\em Given two multisets of values, $M$ and $N$, $M\leq_m N$
iff $occ(M) \leq_{lex} occ(N)$}.
\\\myproof See \cite{kiziltan:MS}.

%\begin{mytheorem}
%Given two vectors $\vec x$ and $\vec y$ of non-empty domains,
%GAC($\vec x \leq_m \vec y$) iff for all $i \in [0,n)$ and $j \in [0,m)$:
%{\small
%\[
%occ(\langle \mymin{x_0},\ldots,\mymax{x_i},\ldots,\mymin{x_{n-1}} \rangle)
%\leq_{lex} occ(\myceiling{\vec y})
%\]
%\[
%occ( \myfloor{\vec x}
%\leq_{lex} occ(\langle \mymax{y_0},\ldots,\mymin{y_j},\ldots,\mymax{y_{m-1}} \rangle)
%\]
%}
%\end{mytheorem}

% ====================================================================
\subsection{A worked example}
\label{sec:WEx}

Based on these lemmas, we have designed an efficient algorithm for
enforcing GAC on a multiset ordering constraint. The algorithm goes
through the $x_i$ and $y_j$ checking for support in the appropriate
occurrence vectors. Incremental computation of the lex ordering test
avoids repetition of work. Consider the multiset ordering constraint
$\vec x \leq_{m}
\vec y$ where:
{\scriptsize \begin{eqnarray*}
\vec{x} &=& \la \{5\}, \{4,5\}, \{3,4,5\}, \{2,4\}, \{1\}, \{1\}\ra\\
\vec{y} &=& \la \{4,5\}, \{4\}, \{1,2,3,4\}, \{2,3\}, \{1\},\{0\}\ra
\end{eqnarray*}}
We construct occurrence vectors for $\myfloor{\vec x}$ and
$\myceiling{\vec y}$, indexed from 5 to 0:
{\scriptsize \begin{eqnarray*}
occ(\myfloor{\vec{x}}) &=& \la 1, 1, 1, 1, 2, 0\ra\\
occ(\myceiling{\vec{y}}) &=& \la 1, 2, 1, 0, 1, 1\ra
\end{eqnarray*}}
Recall that index $i$ in $occ(\myceiling{\vec{y}})$ denotes the number
of occurrences of the value $i$ in $\myceiling{\vec{y}}$. For example,
index 4 is 2 as the value 4 occurs twice.

We first check if $occ(\myfloor{ \vec{x}}) >_{lex}
occ(\myceiling{\vec{y}})$.  If so, we can fail immediately because
no value for any variable can have support.  Here, $occ(\myfloor{
\vec{x}}) \leq_{lex} occ(\myceiling{\vec{y}})$. In fact, we record
(in a pointer, $\alpha$) that the two occurrence vectors are lex
ordered by index 4 of $occ(\myfloor{ \vec{x}})$, which is strictly
smaller than index 4 of $occ(\myceiling{\vec{y}})$.  This means
that we will fail to find support in the $y_j$ if any of the $x_i$
is assigned a new value greater than 4.  We now go through the
$x_i$ checking for support for their maximum values, and then the
$y_i$ checking for support for their minimum values.

Consider $x_0$. As it has a singleton domain, and $occ(\myfloor{
\vec{x}}) \leq_{lex} occ(\myceiling{\vec{y}})$, its only value
must have support so we skip it. Now consider $x_1$. Do its values
have support? Changing $occ( \myfloor{\vec x} )$ to $occ(\langle
\mymin{x_0},\mymax{x_1},\ldots,\mymin{x_{n-1}}
\rangle)$ decreases the number of occurrences of $\mymin{x_1}=4$
by 1, and increases the number of occurrences of $\mymax{x_1}=5$ by
1. As $\mymin{x_1} \geq \alpha =4$, this upsets the lex ordering of
the two occurrence vectors. We therefore prune all values in the
domain of $x_1$ larger than $\alpha$. This leaves a single supported
value, 4. %in the domain of $x_1$.

Now consider $x_2$. Changing $occ(\myfloor{\vec x})$ to
$occ(\langle \mymin{x_0},\ldots,
\mymax{x_2},\ldots,\mymin{x_{n-1}} \rangle)$ decreases the number
of occurrences of $\mymin{x_2}=3$ by 1, and increases the number
of occurrences of $\mymax{x_2}=5$ by 1. As with $x_1$, any value
of $x_2$ larger than $\alpha=4$ upsets the lex ordering.  We
therefore prune $5$ from the domain of $x_2$. Now consider $x_3$.
Changing $occ( \myfloor{\vec x} )$ to $occ(\langle
\mymin{x_0},\ldots,\mymax{x_3},\ldots,\mymin{x_{n-1}} \rangle)$
decreases the number of occurrences of $\mymin{x_3}=2$ by 1, and
increases the number of occurrences of $\mymax{x_3}=4$ by 1.  The
occurrence vectors beneath $\alpha$ would now be lex ordered the
wrong way. We therefore also prune the value $\alpha=4$, leaving a
single supported value 2 in the domain of $x_3$. As $x_4$ and
$x_5$ have singleton domains, their values have support.

Similarly, we check the minimums of the $y_j$ for support. However,
rather than prune values above (and in some cases equal to) $\alpha$,
there is now a dual pointer $\beta$ and we prune values in the domains
of $y_j$ up to (and in some cases equal to) $\beta$. The pointer
$\beta$ is the largest index such that the occurrence vectors beneath
it are lex ordered the wrong way. Any value less than $\beta$ cannot
hope to change the lex ordering as the value at $\beta$ will still
order the vectors the wrong way. Such values can therefore be
pruned. Once we have considered each of the $y_j$, we have the
following generalized arc-consistent vectors: {\scriptsize
\begin{eqnarray*}
\vec{x} &=& \la \{5\}, \{4\}, \{3,4\}, \{2\}, \{1\}, \{1\}\ra\\
\vec{y} &=& \la \{5\}, \{4\}, \{3,4\}, \{2,3\}, \{1\}, \{0\}\ra
\end{eqnarray*}}

\vspace{-2em}

% ====================================================================
\subsection{Algorithm details}

The algorithm uses two pointers $\alpha$ and $\beta$, and two flags
$\gamma$ and $\delta$ to avoid traversing the occurrence vectors each
time we look for support.  The pointer $\alpha$ is set to to the most
significant index above which all occurrences are pair-wise equal and
at $\alpha$ they are strictly ordered.  If the vectors are equal then
$\alpha$ is set to $- \infty$.  The pointer $\beta$ is set to the most
significant index below $\alpha$ such that the occurrence vectors are
lex ordered the wrong way. If no such index exists, we set $\beta$ to
$-\infty$. The flag $\gamma$ is set to true if all the indices between
$\alpha$ and $\beta$ are pair-wise equal and the flag $\delta$ is set
to true if the sub-vectors below $\beta$ are lex ordered the wrong
way. For example, given the occurrence vectors in section
\ref{sec:WEx}, $\alpha$ is set to 4, $\beta$ to 2, and the flags
$\gamma$ and $\delta$ are set to true.

We summarise the major steps the algorithm performs:
%\begin{enumerate}
\begin{description}
%\item Compute $l$ and $u$
\item[A.] Build $occ(\myfloor{\vec x})$ and $occ(\myceiling{\vec y})$
%\item[B.] Fail if  $occ(\myfloor{\vec x}) >_{lex} occ(\myceiling{\vec y})$
\item[B.] Set $\alpha$, $\beta$, $\gamma$, $\delta$ according to their definitions
\item[C.] For each $x_i$:
%\item \hspace{1cm}
If its maximum disturbs the lex ordering on the
occurrence vectors, tighten its upper-bound to $\alpha$ when the
occurrence vectors are lex ordered below $\alpha$,
otherwise to $\alpha -1$.
\item[D.] For each $y_i$:
%\item[E] \hspace{1cm}
If its minimum disturbs the lex ordering on the
occurrence vectors, then tighten its lower-bound to $\beta$ when
the occurrence vectors are lex ordered below
$\alpha$, otherwise to $\beta +1$.
%\end{enumerate}
\end{description}

When we prune a value, we do not need to check recursively for
previous support. Pruning changes neither the lower bounds of $\vec x$
nor the upper bounds of $\vec y$. These values continue to provide
support.  The exception is when a domain is a singleton, and pruning
causes a domain wipe-out.

We now give pseudo-code for an algorithm that maintains GAC on a
multiset ordering constraint between vectors $\vec{x}$ and $\vec{y}$
which are of length $n$ and $m$ respectively. As the algorithm reasons
about occurrences vectors, the original vectors need not be identical
in length (though they often are).

The algorithm is called whenever lower bounds of $x_i$ or upper bounds
of $y_j$ change. Lines {\bf A1} to {\bf A3} build the occurrence
vectors $\vec{ox}$ and $\vec{oy}$. Line {\bf B1} calls the procedure
to set the pointers $\alpha$ and $\beta$, and the flags $\gamma$ and
$\delta$. Lines {\bf C1-13} check support for the maximums of the
$x_i$'s while lines {\bf D1-14} check support for the minimums of the
$y_i$'s.

%\fbox{
%\begin{minipage}[h]
{\scriptsize
%\begin{enumerate} \itemsep = 0pt
\begin{description} \itemsep = 0pt
\item[]{\bf Procedure GACMSO()}
\item[\rm A1.]
%$l:= min (\{\mymin{x_i}| 0 \leq i \leq n-1 \} \cup \{\mymin{y_j}| 0 \leq j \leq m-1 \}) $
$l:= min (\{\mymin{x_i}|  i  \in [0,n) \} \cup \{\mymin{y_j}| j \in [0,m) \}) $
\item[\rm A2.]
%$u:=max (\{\mymax{x_i}| 0 \leq i \leq n-1\} \cup \{\mymax{y_j}| 0 \leq j \leq m-1\}) $
$u:=max (\{\mymax{x_i}| i \in [0,n) \} \cup \{\mymax{y_j}| j \in [0,m) \}) $
\item[\rm A3.] $\vec{ox} = occ(\myfloor{\vec x})$, $\vec{oy} = occ(\myceiling{\vec y})$

\item[\rm B1.] {\bf SetPointers-and-Flags($l, u$)}

\item[\rm C1.] {\bf FOR} $i=0$ {\bf TO} $n-1$~~~~~~ \% check support for x's
\item[\rm C2.] \myhspaceOne {\bf IF} ($\mymin{x_i} \neq \mymax{x_i}$)
\item[\rm C3.] \myhspaceTwo $a:=\mymin{x_i}$ ~~~~~~  $b:=\mymax{x_i}$

\item[\rm C4.] \myhspaceTwo {\bf IF} ($a \geq \alpha$) $NC(x_i=a)$

\item[\rm C5.] \myhspaceTwo {\bf IF} ($\alpha > a > \beta \myAnd b \geq \alpha$)
$NC(x_i \leq \alpha)$

\item[\rm C6.] \myhspaceTwo {\bf IF} ($ a = \beta \myAnd b \geq \alpha$)
\item[\rm C7.] \myhspaceFour {\bf IF} ($ox_{\alpha}+1=oy_{\alpha}$)
\item[\rm C8.] \myhspaceFive {\bf IF} ($\gamma \myAnd ox_{\beta}-1=oy_{\beta} \myAnd \delta$)
$NC(x_i < \alpha)$
\item[\rm C9.] \myhspaceFour {\bf ELSE} $NC(x_i \leq \alpha)$

\item[\rm C10.] \myhspaceTwo {\bf IF} ($ a < \beta \myAnd b \geq \alpha$)
\item[\rm C11.] \myhspaceFour {\bf IF} ($ox_{\alpha}+1 = oy_{\alpha}$)
\item[\rm C12.] \myhspaceFive {\bf IF} ($\gamma$) $NC(x_i < \alpha)$
\item[\rm C13.] \myhspaceFour {\bf ELSE} $NC(x_i \leq \alpha)$

\item[\rm D1.] {\bf FOR} $j=0$ {\bf TO} $m-1$ ~~~~~~\% check support for y's
\item[\rm D2.] \myhspaceOne {\bf IF} ($\mymin{y_j} \neq \mymax{y_j}$)
\item[\rm D3.] \myhspaceTwo $a:=\mymin{y_j}$ ~~~~~ $b:=\mymax{y_j}$
\item[\rm D4.] \myhspaceTwo {\bf IF} ($b > \alpha $)
                            $NC(y_j=b)$
\item[\rm D5.] \myhspaceTwo {\bf IF} ($b = \alpha \myAnd ox_{\alpha}+1 = oy_{\alpha}$)
\item[\rm D6.] \myhspaceThree {\bf IF}($a  \leq \beta \myAnd \gamma$)
\item[\rm D7.] \myhspaceFour {\bf IF} ($ox_{\beta} = oy_{\beta}+1$)
\item[\rm D8.] \myhspaceFive  {\bf IF} ($\delta$)$NC(y_j > \beta)$
\item[\rm D9.] \myhspaceFive {\bf ELSE} $NC(y_j \geq \beta)$
\item[\rm D10.] \myhspaceThree {\bf ELSE} $NC(y_j > \beta)$

%\end{enumerate}
\end{description}
}
%\end{minipage}}

{\scriptsize
%\begin{enumerate} \itemsep = 0pt
\begin{description} \itemsep = 0pt
\item[\rm ] {\bf Procedure SetPointers-and-Flags($l$, $u$)}
%\item \% compute $\alpha$, $\beta$, $\gamma$, $\delta$
\item[\rm B1.] $\gamma$ := {\em false}, $\delta$ := {\em false}, $\alpha := u$
\item[\rm B2.] {\bf WHILE} ($\alpha \geq l \wedge ox_\alpha=oy_\alpha$)
      $\alpha := \alpha - 1$
\item[\rm B3.] {\bf IF} ($\alpha \geq l \wedge ox_\alpha > oy_\alpha$)
      {\bf FAIL}
\item[\rm B4.] {\bf IF} ($\alpha = l-1$) $\alpha := -\infty, \beta := -\infty$
\item[\rm B5.] {\bf ELSE}
\item[\rm B6.] \myhspaceOne $\beta := \alpha - 1$, $\gamma := ${\em true}
\item[\rm B7.] \myhspaceOne {\bf WHILE} ($\beta \geq l \wedge
      ox_\beta \leq oy_\beta$)
\item[\rm B8.] \myhspaceTwo {\bf IF} ($ox_\beta < oy_\beta$) $\gamma :=$ {\em false}
\item[\rm B9.] \myhspaceTwo $\beta := \beta - 1$
\item[\rm B10.] \myhspaceOne {\bf IF} ($\beta = l-1$) $\beta := -\infty$,
      $\gamma :=$ {\em false}
\item[\rm B11.] {\bf IF} ($\beta \neq -\infty \wedge \beta > l$)
\item[\rm B12.] \myhspaceOne $i := \beta - 1$
\item[\rm B13.] \myhspaceOne {\bf WHILE} ($i \geq l \wedge ox_i = oy_i$)
      $i := i - 1$
\item[\rm B14.] \myhspaceOne {\bf IF} ($i \geq l \wedge ox_i > oy_i$)
      $\delta := $ {\em true}
%\end{enumerate}
\end{description}

}

For each $x_i$, we only check for support if $x_i$ is not singleton
(line {\bf C2}). There are six cases where we prune the domain of
$x_i$: (1) at line {\bf C4}, as $\mymin{x_i} \geq \alpha$, any value in
the domain of $x_i$ greater than $\mymin{x_i}$ lacks support because
it will disturb the lex ordering; (2) at line {\bf C5}, as $\alpha >
\mymin{x_i} > \beta$, and $\mymax{x_i} \geq \alpha$, all the values in
the domain of $x_i$ greater than $\alpha$ disturb the lex ordering,
and lack support; (3) at line {\bf C6}, {\bf C7}, and {\bf C8}, all
values greater than or equal to $\alpha$ lack support.  If we assign
$\alpha$ to $x_i$, then the vectors will be equal at index $\alpha$
and $\beta$, the values between $\alpha$ and $\beta$ are also all
pair-wise equal (since $\gamma$ is true), and the vectors below
$\beta$ are ordered the wrong way (since $\delta$ is true). Thus, the
value $\alpha$ also lacks support and is pruned; (4) at line {\bf C9},
all values greater than $\alpha$ lack support; (5) at line {\bf C10},
{\bf C11}, and {\bf C12}, all values greater than or equal to $\alpha$
lack support. If we assign $\alpha$ to $x_i$, the vectors will be lex
ordered the wrong way as the difference between the number of
occurrences at $\alpha$ is exactly one, and $\gamma$ is true. Thus,
the value $\alpha$ also lacks support and is pruned; (6) at line {\bf
C13} all values greater than $\alpha$ lack support.

For each $y_j$, we only check for support if $y_j$ is not singleton
(line {\bf D2}). There are four cases where we prune the domain of
$y_j$: (1) at line {\bf D4}, as $\mymax{y_j} > \alpha$, any value
smaller than $\mymax{y_j}$ disturbs the lex ordering and lacks
support; (2) at lines {\bf D5} to {\bf D8}, the situation is dual to
the third case for the $x_i$, and any value less than or equal to $\beta$
lacks support; (3) at line {\bf D9}, any value less than $\beta$ lacks
support; (4) at line {\bf D10}, again, any value less than or equal to
$\beta$ lacks support.

\subsection{Theoretical properties}

In a longer technical report, we prove
the following results about the algorithm's complexity and correctness. 

\begin{mytheorem}
{\bf GACMSO} runs in time $O(n+m+d)$ where $d$ is $u-l$. 
\end{mytheorem}
If
$d \ll n, m$ (and for multisets, we expect
this as the number of values is typically less
than the cardinalities to permit repetition),
then the algorithm is $O(n+m)$.

\begin{mytheorem}
For disjoint and non-repeating vectors,
{\bf GACMSO} either establishes failure if $\vec x \leq_m \vec
y$ is not satisfiable, or prunes values from $\vec x$ and $\vec y$ to
ensure $GAC(\vec x \leq_m \vec y)$.
\end{mytheorem}

The algorithm can easily be modified to support strict multiset
ordering. The only differences are that we fail if $\vec{ox}=\vec{oy}$
and that $\beta$ is set to $l-1$ not $-
\infty$. The algorithm then enforces a strict inequality on the
occurrence sub-vectors above $\beta$. Another variant of the algorithm
is when $d \gg n, m$. In such a situation, it could be costly to
construct the occurrence vectors. We can instead sort the minimums of
the $x_i$ and the maximums of the $y_j$, and compute 
$\alpha$, $\beta$, $\gamma$ and $\delta$ as if
we had the occurrences by scanning these sorted lists. This
information is all we need to compute support for each
$x_i$ and $y_j$ in turn. The complexity of this
modified algorithm is $O(n \log n + m \log m)$ as the cost of
sorting dominates.

If we have multiple rows of a matrix that are multiset ordered,
we can decompose this into multiset ordering constraint
on all pairs of rows, or (further still) onto ordering
constraints just on neighbouring pairs of rows. 
The following result shows that such decompositions 
hinder constraint propagation. Nevertheless, 
it will usually be
most cost effective to post just the $O(n)$ ordering
constraints between neighbouring pairs
rather than the $O(n^2)$ constraints between all
pairs. 
\begin{mytheorem}
$GAC(\forall i < j \ . \ \vec{x}_i \leq_{m} \vec{x}_j) $
is strictly stronger than $GAC( \vec{x}_i \leq_{m} \vec{x}_j) $
for all $i < j$, and this itself is
strictly stronger than
$GAC( \vec{x}_i \leq_{m} \vec{x}_{i+1}) $
for all $i$.
\end{mytheorem}
%\myproof
%Consider the following 3 vectors:
%$
%\vec{x}_0 = \langle \{ 0, 1\}, \{ 1 \}, \{ 0, 1\} \rangle$,
%$\vec{x}_1 = \langle \{ 0, 1\}, \{ 0, 1 \}, \{ 0, 1\} \rangle$,
%$\vec{x}_2 = \langle \{ 0, 1\}, \{ 0 \}, \{ 0\} \rangle$.
%Although $GAC(\vec{x}_i \leq_{m} \vec{x}_{i+1})$ for
%all $i$, $GAC(\vec{x}_0 \leq_{m} \vec{x}_{2})$ does
%not hold as the value $x_{0,0} = \{1\}$ cannot be
%consistently extended.
%Now, replace $\vec{x}_2$ by
%$\vec{x}_2 = \langle \{ 0, 1\}, \{ 0 \}, \{ 0, 1\} \rangle$.
%Then $GAC(\vec{x}_i <_{m} \vec{x}_{i+1})$ for
%all $i$, $GAC(\vec{x}_0 <_{m} \vec{x}_{2})$ does
%not hold as the value $x_{0,0} = \{1\}$ cannot be
%consistently extended.
%\myqed
%
%With vectors of 0/1 variables, lex ordering all pairs
%is equivalent to global consistency, but
%strict lex ordering all pairs is
%weaker than global consistency.
%These results map over to multiset orderings.
%However, they are much less interesting
%as a multiset ordering of vectors of 0/1 variables
%is equivalent simply to ordering their sums.

\section{Alternative approaches}

\subsection{Arithmetic constraint}

Barbara Smith (personal communication) has proposed enforcing $\vec x
<_{m} \vec y$ on vectors of length $n$ via the arithmetic constraint $
n^{x_0} + \ldots n^{x_{n-1}} < n^{y_0} + \ldots n^{y_{n-1}} $. This is
similar to the transformation of a leximin fuzzy CSP into an
equivalent MAX CSP
\cite{sfvijcai95}. %It differs because, as opposed to the fixed
%satisfaction degrees associated with the tuples of a fuzzy constraint,
%the values assigned to the variables are not known {\em a priori}. 
BC on such a constraint is equivalent to GAC on the original multiset
ordering constraint. However, such an arithmetic constraint is only
feasible for small $n$. Further, most existing solvers will not
enforce BC on such an arithmetic constraint, but will delay it until
all but one of the variables are instantiated.

%\subsection{Decomposition via occurrence vectors}
\subsection{Decomposition}

Multiset ordering is equivalent to the lex ordering the associated
occurrence vectors. As we have efficient algorithms for
constructing occurrence vectors (via the global cardinality
constraint \cite{regin2}) and for lex ordering \cite{fhkmwcp2002},
this might be an alternative approach.
% for posting
%multiset ordering constraints.
%We can channel to an occurrence vector
%using Regin's global cardinality constraint \cite{regin2},
%and then lex order these.
However, as the following theorem shows, such a decomposition
hinders constraint propagation. Also, the two global cardinality
constraints in such a decomposition
are more expensive to enforce than the %$O(n+m)$
algorithm presented here. We write $gcc(\vec x,\vec{ox})$ for the
global cardinality constraint that channels from a vector of
variables $\vec x$ to the associated occurrence vector $\vec{ox}$.

\begin{mytheorem}
$GAC(\vec x <_m \vec y)$ is
strictly stronger than
simultaneously enforcing $GAC(gcc(\vec x,\vec{ox}))$,
$GAC(gcc(\vec y,\vec{oy}))$, and
$GAC(\vec{ox} <_{lex} \vec{oy})$.
\end{mytheorem}
\myproof Clearly it is as strong. To show strictness, consider
$\vec x  =  \la \{ 1, 2 \}, \{ 1, 2 \}, \{ 2\}, \{2\} \ra$ and $
\vec y  =  \la \{ 1, 2 \}, \{ 1, 2 \}, \{ 0, 1, 2\}, \{ 0, 1\}
\ra$.
%ZEYNEP
%\begin{eqnarray*}
%\vec x & = & \la \{ 1, 2 \}, \{ 1, 2 \}, \{ 2\}, \{2\} \ra \\
%\vec y & = & \la \{ 1, 2 \}, \{ 1, 2 \}, \{ 0, 1, 2\}, \{ 0, 1\} \ra
%\end{eqnarray*}
The multiset ordering constraint
is not GAC since 0 in $y_2$ has no support but the decomposition
is unable to determine this.
% ZEYNEP
%This is not GAC since $y_2$ needs to be pruned to give:
%\begin{eqnarray*}
%\vec y & = & \la \{ 1, 2 \}, \{ 1, 2 \}, \{ 1, 2\}, \{ 0, 1\} \ra
%\end{eqnarray*}
%The corresponding occurrence vectors are
%$\vec{ox}  =  \langle \{0\}, \{0, 1, 2\} ,\{2, 3, 4\} \rangle$ and
%$\vec{ox}  =  \langle  \{2, 3, 4\}, \{0, 1, 2\}, \{0\}\rangle$ and
%$\vec{oy}  = \langle \{0, 1, 2\}, \{0, 1, 2, 3, 4\} ,\{0, 1, 2, 3\} \rangle$.
%$\vec{oy}  = \langle  \{0, 1, 2, 3\}, \{0, 1, 2, 3, 4\}, \{0, 1, 2\} \rangle$.
%ZEYNEP
%\begin{eqnarray*}
%\vec{ox} & = & \langle \{0\}, \{0, 1, 2\} ,\{2, 3, 4\} \rangle \\
%\vec{oy} & = & \langle \{0, 1, 2\}, \{0, 1, 2, 3, 4\} ,\{0, 1, 2, 3\}  \rangle
%\end{eqnarray*}
%Enforcing GAC on the lex ordering constraint gives
%$\vec{ox}  = \langle \{0\}, \{0, 1, 2\} ,\{2, 3\} \rangle$ and
%$\vec{ox}  = \langle \{2, 3\}, \{0, 1, 2\}, \{0\} \rangle$ and
%$\vec{oy}  = \langle \{0, 1, 2\}, \{0, 1, 2, 3, 4\} ,\{2, 3\} \rangle$.
%$\vec{oy}  = \langle \{2, 3\}, \{0, 1, 2, 3, 4\}, \{0, 1, 2\} \rangle$.
%ZEYNEP
%\begin{eqnarray*}
%\vec{ox} & = & \langle \{0\}, \{0, 1, 2\} ,\{2, 3\} \rangle \\
%\vec{oy} & = & \langle \{0, 1, 2\}, \{0, 1, 2, 3, 4\} ,\{2, 3\}  \rangle
%\end{eqnarray*}
%But, this does not allow us to prune any values from $\vec y$.
\myqed

Another approach is to use the sorted constraint in the Eclipse
solver. This ensures that the values taken by one vector of variables
are identical but in sorted order to the values taken by a second
vector of variables. To post a multiset ordering constraint on two
vectors, we can channel each into a sorted vector and lex order these.
The above example demonstrates that such a decomposition again hinders
propagation. The sorting constraint is also more expensive to enforce.

% than the %$O(n+m)$
%GAC algorithm presented here.

%ZEYNEP
%\subsection{0/1 multisets}
%
%One special case that often concerns us is ordering 0/1 vectors.
%Multiset ordering then
%reduces to ordering the vector sums. As bounds consistency
%techniques will reason about such sums quickly and effectively, we
%need not consider multiset orderings when dealing with 0/1
%vectors, but can instead deal with the sums directly.

\section{Experimental results}

We designed some experiments to test three goals.
First, is multiset ordering an
effective method for dealing with row and/or column symmetry?
Second, how does multiset ordering compare to lex ordering? Which one
breaks more symmetry? Is a combined method, which multiset orders
one dimension and lex orders the other one of the matrix, superior?
Third, does our GAC algorithm do more
inference in practice than the decomposition? Similarly, is the
algorithm more efficient in practice than its decomposition?

The symmetry breaking constraints we used are strict
lex ordering on the columns ($<_{lex}$C), on the rows
($<_{lex}$R); multiset ordering on the rows ($\leq_{m}$R),
(strict) multiset ordering on the columns ($\leq_{m}$C and
$<_{m}$C); and combinations of these constraints.
%($<_{lex}$RC,
%$\leq_{m}$RC, $<_{lex}$C $\leq_{m}$R, $\leq_{m}$C $<_{lex}$R).
Such constraints are posted between adjacent
rows/columns.
The results of the experiments are shown in tables where a ``-'' means
no result is obtained in 1 hour (3600 secs). The experiments are done
using ILOG Solver 5.2 on a 1000MHz pentimum III with 256 Mb RAM.

\subsection{Progressive Party Problem}
%In the progressive party problem (prob013 in CSPLib), we
There are a set of host boats, each with a capacity, and a set of
guest boats, each with a crew size. We wish to assign a host for
each guest and period, such that a guest crew never visits the
same host twice, no two guest crews meet more than once, and the
spare capacity of each host boat, after accommodating its own
crew, is not exceeded (prob013 in csplib.org).

A matrix model of this problem \cite{smith95:PPP} is a 2-d matrix
of $guests \times periods$ where each variable is assigned a host
representing that a host is accommodating a particular guest in a
given time period. The rows are the guests, the columns are the
periods. This model has column and partial row symmetry: any two
periods, and any two guests with the same crew size are
indistinguishable.
%We have experimented with this model by considering
We consider the 13-hosts and 29 guests problem  with 5 and 6 time
periods, referred as 5-13-29 and 6-13-29. These problems have
$p!14!2!4!5!7!$ row and column symmetries where $p$ is the number of
time periods. The
actual data can be found in csplib.org. Due to the problem
constraints, no pair of rows/columns can be equal, hence we can
safely pose strict lex ordering. However, any two distinct
rows/columns might be equal when viewed as multisets.

\begin{table}[t]
\begin{scriptsize}
\begin{center}
\begin{tabular}{|r||r|r|r||}
\hline
%\hline & \multicolumn{3} {|c|} {Problem} \\
%\hline Model &\multicolumn{3}{|c|}{5-13-29}\\
\hline &Fails&Choice points &Time (secs.)  \\
\hline No Symmetry Breaking &180,738& 180,860& 75.9  \\
\hline $<_{lex}$C &180,738 & 180,860 & 81.5  \\
\hline $<_{lex}$R &2,720 & 2,842& 2.7 \\
\hline $<_{lex}$RC &2,720 & 2,842& 2.7 \\
\hline $\leq_{m}$C &137,185 & 137,306& 71.2 \\
%\hline $gcc+\leq_{lex}$C &142,235 & 142,356& 74.517& -& -&- \\
\hline $\leq_{m}$R &10,853 & 10,977& 8.6\\
%\hline $gcc+\leq_{lex}$R &20,367 & 20,491& 10.525& -& -&- \\
\hline $\leq_{m}$RC &- & -& - \\
\hline $<_{lex}$C $\leq_{m}$R &10,853 & 10,977& 8.6 \\
\hline $\leq_{m}$C $<_{lex}$R  &2,016 & 2,137& 2.6 \\
%\hline $gcc+\leq_{lex}$C $<_{lex}$R  &2,030 & 2,151& 1.992& -& -&- \\
\hline
\end{tabular}
\end{center}
\caption{\label{table:PPPResults1}
5-13-29 progressive party problem using row-by-row labelling. } \end{scriptsize}
\end{table}

As in \cite{smith95:PPP}, the guest boats are ordered in descending
order of their size. We order the host boats in descending order of
spare capacity to choose a value in a succeed-first manner.  Results
obtained by row-by-row, and column-by-column labelling strategies are
given in Tables \ref{table:PPPResults1} and
\ref{table:PPPResults2}. With row-by-row labelling, we cannot solve 6-13-29 with or without symmetry breaking. For
the 5-13-29 problem, $<_{lex}$R breaks a lot more row symmetry than
$\leq_{m}$R. However, the reverse is true for the columns. Here,
$<_{lex}$C does not break any symmetry but $\leq_{m}$C does.  Multiset
ordering one dimension of a matrix therefore does not necessarily
break less symmetry than lex ordering the same dimension. Such
phenomena occur through interactions with the search strategy: a
search strategy might already lex order, so multiset ordering
constraints break additional symmetry.  The smallest search tree and
also the least solving time is obtained by $\leq_{m}$C
$<_{lex}$R. This supports our conjecture that lex ordering one
dimension combined with multiset ordering the other can break more
symmetry than lex/multiset ordering both dimensions.

\begin{table}[t]
\begin{scriptsize}
\begin{center}
\begin{tabular}{|r||r|r|r||r|r|r||}
\hline
%\hline & \multicolumn{3} {|c|} {Problem} \\
%\hline Model & \multicolumn{3}{|c|}{6-13-29}\\
\hline &Fails&Choice points & Time (secs.) \\
\hline No Symmetry Breaking & 20,722& 20,871&12.3 \\
\hline $<_{lex}$C &20,722& 20,871&12.4 \\
\hline $<_{lex}$R &20,722& 20,871&12.5 \\
\hline $<_{lex}$RC &20,722& 20,871&12.4 \\
\hline $\leq_{m}$C & 7,053 & 7,202 & 4.6 \\
%\hline $gcc+\leq_{lex}$C &8,387 & 8,517& 2.844 & 8,414 & 8,563 & 3.885 \\
\hline $\leq_{m}$R & -& -&- \\
\hline $\leq_{m}$RC & -& -&- \\
\hline $<_{lex}$C $\leq_{m}$R & -& -&- \\
\hline $\leq_{m}$C $<_{lex}$R  & 7,053 & 7,202 & 4.6 \\
\hline
\end{tabular}
\end{center}
\caption{\label{table:PPPResults2} 6-13-29 progressive party problem
using column-by-column labelling. }
\end{scriptsize}
\end{table}

%With the other labelling strategies, we obtain the same search
%tree.
%In Table \ref{table:PPPResults2}, we give the run-times
%obtained by column-by-column labelling, though the other two
%labellings solve the problems in similar time units.
%column-by-column labellingWe are now
With column-by-column labelling, we are able to solve the 6-13-29
problem. Neither of $<_{lex}$R, $<_{lex}$C, $<_{lex}$RC break any
symmetry. The smallest search tree is obtained by $\leq_{m}$C.
This supports our conjecture that multiset ordering one dimension
can break more symmetry than lex ordering the same or both
dimensions. If the search strategy already orders both dimensions
lexicographically, imposing a constraint like multiset ordering in
one dimension breaks additional symmetry.

\subsection{Sports Scheduling with Odd Teams}
This is a modified version of prob026 in csplib.org. We have $n$
teams ($n$ is odd), playing over $n$ weeks. Each week is divided
into $(n-1)/2$ periods, and each period is divided into 2 slots,
home and away.
%The team in the first slot plays at home against the team in the second.
We wish to find a schedule so that every team plays at
most once a week, every team plays twice in the same period over
the tournament and every team plays every other team.  We slightly
modify the model in \cite{vanhentenryck1}, where {\em teams} is a
3-d matrix of $periods \times weeks \times slots$.  Each element
of {\em teams} is the team playing in a given period, week and
slot. We treat this matrix as 2-d where the rows are the periods
and columns are the weeks, and each entry is a list of variables
giving the slots.

As the periods and the weeks are indistinguishable, this problem
has $n!(n-1/2)!$ row and column symmetries. We pose strict ordering constraints
on the rows and columns of $teams$ as the periods and weeks cannot
be equal. Due to the constraints on the periods, posing multiset
ordering on the rows is not effective.

Results obtained by column-by-column labelling of the $teams$ are
given in Table \ref{table:OddSportsResults3-1}. For one column, we
first label the first slots; for the other, we first label the
second slots. With this strategy, $<_{lex}$R does not break any
symmetry, so we omit it in the table. Posing multiset ordering by
our algorithm is much more effective and efficient than by $gcc$
and lex ordering constraints. This holds for many other search
strategies. In Table \ref{table:OddSportsResults3-1}, we note that
$<_{m}$C gives a smaller search tree than $<_{lex}$C. However, for
other search strategies the reverse is true. This supports the
theoretical result that lex ordering and multiset ordering are
incomparable.

\section{Conclusions}

We have identified a new and important global (non-binary)
constraint. This constraint ensures that the values taken by two
vectors of variables, when viewed as multisets, are ordered. We
have developed an efficient linear time algorithm for enforcing
generalised arc-consistency on such a multiset ordering
constraint. We have proposed a number of applications for this new
constraint including breaking symmetry in matrix models, and fuzzy
constraint satisfaction. We have shown that alternative methods
for posting a multiset ordering constraint like an arithmetic
constraint or decomposition are inferior. Finally, we have
implemented this generalized arc-consistency algorithm in ILOG
Solver. Experimental results on a number of problem domains show
considerable promise.

\begin{table}[t]
\begin{scriptsize}
\begin{center}
\begin{tabular}{|c|r|r|r|r|l|}
\hline $n$ & Model & Failures & Choice points & Time (sec.) \\
\hline
$5$ &No symmetry breaking & 3 & 12 & 1.1  \\

    & $gcc+<_{lex}$C & 2 & 11& 1.4  \\

    &$<_{m}$C & 1 & 10 & 1.4  \\

    &$<_{lex}$C &  3& 12& 1.5 \\

\hline

$7$ &No symmetry breaking &6,871 & 6,890 & 1.9  \\

    & $gcc+<_{lex}$C & 74 & 92 & 1.3  \\

    &$<_{m}$C & 69 & 87 & 1.1  \\

    &$<_{lex}$C &  771 & 788 & 1.3 \\

\hline

$9$ &No symmetry breaking &- & - & -  \\

    & $gcc+<_{lex}$C & 2,616,149 & 2,616,177 & 857.2  \\

    &$<_{m}$C & 760,973 & 761,003 & 130.5  \\

    &$<_{lex}$C &  - & - & - \\

\hline

\end{tabular}
\end{center} \caption{\label{table:OddSportsResults3-1} Sports scheduling problem.}
\end{scriptsize}
\end{table}

\ifx\printing\article \vspace{-2mm}
\section*{Acknowledgements}
This research is supported by EPSRC under GR/R30792 and by the
Science Foundation Ireland.  We thank the other members of the 4C
lab (especially Chris Beck) and APES research group, and Chris
Jefferson. \fi

\bibliographystyle{named}
%\bibliographystyle{alpha}
%\bibliography{/home/s5/tw/biblio/a-z,/home/s5/tw/biblio/pub}
%\bibliography{/n/endjinn/u6/tw/biblio/a-z,/n/endjinn/u6/tw/biblio/pub}
%\bibliography{/usr/tw/biblio/a-z,/usr/tw/biblio/pub}
%\bibliography{/home/arp/disk1/tw/biblio/a-z,/home/arp/disk1/tw/biblio/pub}
%\bibliography{/u6/tw/biblio/a-z,/u6/tw/biblio/pub}
%\bibliography{/usr/local/users/tw/biblio/a-z,/usr/local/users/tw/biblio/pub}

\begin{thebibliography}{} \itemsep=-1pt

\bibitem[\protect\citeauthoryear{Beldiceanu}{2000}]{beldiceanu1}
N.~Beldiceanu.
\newblock Global constraints as graph properties on a structured network of
  elementary constraints of the same type.
\newblock In {\em Proc. of CP-2000}, LNCS 1894, pp 52--66, 2000.

%\bibitem[\protect\citeauthoryear{Beldiceanu and Contegean}{1994}]{beldiceanu2}
%N.~Beldiceanu and E.~Contegean.
%\newblock Introducing global constraints in {CHIP}.
%\newblock {\em Mathematical Computer Modelling}, 20(12):97--123, 1994.

%\bibitem[\protect\citeauthoryear{Bessiere and
%  R\'{e}gin}{1997}]{Bessiere-Regin97}
%C.~Bessiere and J.C. R\'{e}gin.
%\newblock Arc consistency for general constraint networks: Preliminary results.
%\newblock In {\em Proc. of the 15th IJCAI},
%  pages 398--404. %International Joint Conference on Artificial Intelligence,
%  1997.

\bibitem[\protect\citeauthoryear{Fahle \bgroup \em et al.\egroup
  }{2001}]{fahle:SBDD}
T.~Fahle, S.~Schamberger and M.~Sellman.
\newblock Symmetry Breaking.
\newblock In {\em Proc. of CP-2001}, LNCS 2239, pp 93--107, 2001.

\bibitem[\protect\citeauthoryear{Fargier}{1994}]{fargier:Thesis}
H.~Fargier.
\newblock {\em Probl\`emes de satisfaction de constraintes flexibles:
  application \`a l'ordonnancement de production}.
\newblock PhD thesis, Uni. of Paul Sabatier, Toulouse, 1994.

\bibitem[\protect\citeauthoryear{Flener \bgroup \em et al.\egroup
  }{2002}]{ffhkmpwcp2002}
P.~Flener, A.~Frisch, B.~Hnich, Z.~Kiziltan, I.~Miguel,
J.~Pearson, and
  T.~Walsh.
\newblock Breaking row and column symmetry in matrix models.
\newblock In {\em Proc. of CP-2002}, LNCS 2470, pp 462--476, 2002.

\bibitem[\protect\citeauthoryear{Frisch \bgroup \em et al.\egroup
  }{2002}]{fhkmwcp2002} A.~Frisch, B.~Hnich, Z.~Kiziltan, I.~Miguel,
and T.~Walsh.
\newblock Global constraints for lexicographic orderings.
\newblock In {\em Proc. of CP-2002}, LNCS 2470, pp 93--108, 2002.

%\bibitem[\protect\citeauthoryear{Gent and Walsh}{1999}]{csplib}
%I.P. Gent and T.~Walsh.
%\newblock CSPLib: a benchmark library for constraints.
%%\newblock Technical report, Technical report APES-09-1999, 1999.
%%\newblock Available from \verb+http://dcs.st-and.ac.uk/~apes+.
%%A shorter version appears in the Proceedings of the 5th International Conference on
%%  Principles and Practices of Constraint Programming (CP-99).
%\newblock In {\em 5th Int. Conf. on Principles and Practices of
%  Constraint Programming (CP-99)}. Springer, 1999.

\bibitem[\protect\citeauthoryear{Gent and Smith}{2000}]{gent:SBDS}
I.~Gent and B.~Smith.
\newblock Symmetry Breaking in Constraint Programming.
\newblock In {\em Proc. of ECAI '2000}, pp 599--603, 2000.

\bibitem[\protect\citeauthoryear{Kiziltan and Smith}{2002}]{kscp02}
Z.~Kiziltan and B.~Smith.
\newblock Symmetry breaking constraints for matrix models.
\newblock In {\em Proc. of 2nd Int. Workshop on Symmetry in
  Constraint Satisfaction Problems (SymCon-02)}, held alongside CP-2002, 2002.

\bibitem[\protect\citeauthoryear{Kiziltan and Walsh}{2002}]{kiziltan:MS}
Z.~Kiziltan and T.~Walsh.
\newblock Constraint Programming with Multisets.
\newblock In {\em Proc. of 2nd Int. Workshop on Symmetry in
  Constraint Satisfaction Problems (SymCon-02)}, held alongside
  CP-2002, 2002.

%\bibitem[\protect\citeauthoryear{R\'{e}gin and Rueher}{2000}]{regin4}
%J.C. R\'{e}gin and M.~Rueher.
%\newblock A global constraint combining a sum constraint and difference
%  constraints.
%\newblock In
%% R.~Dechter, editor,
%%{\em Proceedings of
%{\em 6th Int. Conf. on Principles and Practice of Constraint Programming (CP2000)},
%  pages 384--395. Springer, 2000.

\bibitem[\protect\citeauthoryear{R\'{e}gin}{1994}]{regin1}
J.C. R\'{e}gin.
\newblock A filtering algorithm for constraints of difference in {CSPs}.
\newblock In {\em Proc. AAAI'94}, pp
  362--367. %American Association for Artificial Intelligence,
1994.

\bibitem[\protect\citeauthoryear{R\'{e}gin}{1996}]{regin2}
J.C. R\'{e}gin.
\newblock Generalized arc consistency for global cardinality constraints.
\newblock In {\em Proc. of AAAI'96}, pp
  209--215. %American Association for Artificial Intelligence, AAAI Press/The
%  MIT Press,
1996.

\bibitem[\protect\citeauthoryear{Schiex \bgroup \em et al.\egroup
  }{1995}]{sfvijcai95}
T.~Schiex, H.~Fargier, and G.~Verfaille.
\newblock Valued constraint satisfaction problems: hard and easy
problems.
\newblock In {\em Proc. of IJCAI'95}, pp 631--637. 1995.

\bibitem[\protect\citeauthoryear{Smith \bgroup \em et al.\egroup
  }{1995}]{smith95:PPP} B.M. Smith, S.C. Brailsford,
P.M. Hubbard, and H.P. Williams.
\newblock The progressive party
%problem: integer linear programming and constraint programming
%compared. 
problem. \newblock Research Report 95.8, Uni. of Leeds. 1995.

\bibitem[\protect\citeauthoryear{Van Hentenryck \bgroup \em et al.\egroup
  }{1999}] {vanhentenryck1}
P.~van Hentenryck, L.~Michel, L.~Perron, and J.C. R\'{e}gin.
\newblock Constraint programming in {OPL}. \newblock In {\em Proc.
of PPDP'99}, pp 98--116. Springer, 1999.

\end{thebibliography}

\end{document}